\def\year{2022}\relax
\documentclass[letterpaper]{article} 
\usepackage{aaai22}  
\usepackage{times}  
\usepackage{helvet}  
\usepackage{courier}  
\usepackage[hyphens]{url}  
\usepackage{graphicx} 
\usepackage{natbib}  
\usepackage{caption} 
\DeclareCaptionStyle{ruled}{labelfont=normalfont,labelsep=colon,strut=off} 
\usepackage{algorithm}
\usepackage{algorithmic}
\usepackage{cite}
\usepackage{amsmath,amssymb,amsfonts}
\usepackage{algorithmic}
\usepackage{graphicx}
\usepackage{textcomp}
\usepackage{xcolor}
\usepackage{soul}
\usepackage{tikz}
\usetikzlibrary{automata}
\usetikzlibrary{calc, automata, chains, arrows.meta}
\usetikzlibrary{chains,scopes}
\usetikzlibrary{positioning}
\usepackage{caption}
\usepackage{subcaption}
\usepackage{multirow}
\usepackage{mathtools}
\usepackage{tablefootnote}
\usepackage{url}
\usepackage{float}

\usepackage{newfloat}
\usepackage{listings}
\floatstyle{ruled}
\newfloat{listing}{tb}{lst}{}
\floatname{listing}{Listing}
\nocopyright
\def\BibTeX{{\rm B\kern-.05em{\sc i\kern-.025em b}\kern-.08em
    T\kern-.1667em\lower.7ex\hbox{E}\kern-.125emX}}

\title{Field Study in Deploying Restless Multi-Armed Bandits: Assisting Non-Profits in Improving Maternal and Child Health
}

\author{
    Aditya Mate\textsuperscript{\rm 1}\textsuperscript{\rm 2}\thanks{Work done during an internship at Google Research}\equalcontrib,
    Lovish Madaan\textsuperscript{\rm 1}\equalcontrib,
    Aparna Taneja\textsuperscript{\rm 1},
    Neha Madhiwalla\textsuperscript{\rm 3},
    Shresth Verma\textsuperscript{\rm 1},
    \\
    Gargi Singh\textsuperscript{\rm 1},
    Aparna Hegde\textsuperscript{\rm 3},
    Pradeep Varakantham\textsuperscript{\rm 4},
    Milind Tambe\textsuperscript{\rm 1}
}
\affiliations{
    \textsuperscript{\rm 1} Google Research India\\
    \textsuperscript{\rm 2} Harvard University\\
    \textsuperscript{\rm 3} ARMMAN\\
    \textsuperscript{\rm 4} Singapore Management University\\

}

\usepackage{bibentry}

\begin{document}

\maketitle

\begin{abstract}
The widespread availability of cell phones has enabled non-profits to deliver critical health information to their beneficiaries in a timely manner.
This paper describes our work to assist non-profits that employ automated messaging programs to deliver timely preventive care information to beneficiaries (new and expecting mothers) during pregnancy and after delivery. Unfortunately, a key challenge in such information delivery programs is that a significant fraction of beneficiaries drop out of the program. Yet, non-profits often have limited health-worker resources (time) to place crucial service calls for live interaction with beneficiaries to prevent such engagement drops. To assist non-profits in optimizing this limited resource, we developed a Restless Multi-Armed Bandits (RMABs) system. One key technical contribution in this system is a novel clustering method of offline historical data to infer unknown RMAB parameters. Our second major contribution is evaluation of our RMAB system in collaboration with an NGO, via a real-world service quality improvement study.
The study compared strategies for optimizing service calls to 23003 participants over a period of 7 weeks to reduce engagement drops. We show that the RMAB group provides statistically significant improvement over other comparison groups, reducing $\sim30\%$ engagement drops. To the best of our knowledge, this is the first study demonstrating the utility of RMABs in real world public health settings. We are transitioning our RMAB system to the NGO for real-world use.

\end{abstract}

\section{Introduction}
\label{sec:intro}
The wide-spread availability of cell phones has allowed non-profits to deliver targeted health information via voice or text messages to beneficiaries in underserved communities, often with significant demonstrated benefits to those communities \cite{pfammatter2016mhealth, Kaur2020smart}.
We focus in particular on non-profits that target improving maternal and infant health in low-resource communities in the global south. These non-profits deliver ante- and post-natal care information via voice and text to prevent adverse health outcomes \cite{mom-connect, mMitra, helpmum}. 

Unfortunately, such information delivery programs are often faced with a key shortcoming: a large fraction of beneficiaries who enroll may drop out or reduce engagement with the information program. Yet non-profits often have limited health-worker time available on a periodic (weekly) basis to help prevent engagement drops. More specifically, there is limited availability of health-worker time where they can place crucial service calls (phone calls) to a limited number of beneficiaries, to encourage beneficiaries' participation, address complaints and thus prevent engagement drops. 

Optimizing limited health worker resources to prevent engagement drops requires that we prioritize beneficiaries who would benefit most from service calls on a periodic (e.g., weekly) basis. We model this resource optimization problem using Restless Multi-Armed Bandits (RMABs), with each beneficiary modeled as an RMAB arm. RMABs have been well studied for allocation of limited resources motivated by a myriad of application domains including preventive interventions for healthcare \cite{mate2020collapsing}, planning anti-poaching patrols \cite{QianRestlessPoachers}, machine repair and sensor maintenance \cite{glazebrook2006indexable} and communication systems \cite{sombabu_paper}. However, RMABs have rarely seen real world deployment, and to the best of our knowledge, never been deployed in the context of large-scale public health applications. 

This paper presents first results of an RMAB system in real world public health settings. Based on available health worker time, RMABs choose $m$ out of $N$ total beneficiaries on a periodic (e.g., weekly) basis for service calls, where the $m$ are chosen to optimize prevention of engagement drops. The paper presents two main contributions. First, previous work often assumes RMAB parameters as either known or easily learned over long periods of deployment. We show that both assumptions do not hold in our real-world contexts; instead, we present clustering of offline historical data as a novel approach to infer unknown RMAB parameters. 

Our second contribution is a real world evaluation showing the benefit of our RMAB system, conducted in partnership with ARMMAN\footnote{https://armman.org/}, an NGO in India focused on maternal and child care. ARMMAN conducts a large-scale health information program, with concrete evidence of health benefits, which has so far served over a million mothers. As part of this program, an automated voice message is delivered to an expecting or new mother (beneficiary) over her cell phone on a weekly basis throughout pregnancy and for a year post birth in a language and time slot of her preference. 

Unfortunately, ARMMAN's information delivery program also suffers from engagement drops.
Therefore, in collaboration with ARMMAN we conducted a service quality improvement study to maximize the effectiveness of their service calls to ensure beneficiaries do not drop off from the program or stop listening to weekly voice messages. More specifically, the current standard of care in ARMMAN's program is that any beneficiary may initiate a service call by placing a so called ``missed call''. This beneficiary-initiated service call is intended to help address beneficiaries' complaints and requests, thus encouraging engagement. However, given the overall decreasing engagement numbers in the current setup, key questions for our study are to investigate an approach for effectively conducting additional ARMMAN-initiated service calls (these are limited in number) to reduce engagement drops. To that end, our service quality improvement study comprised of 23,003 real-world beneficiaries spanning $7$ weeks. 
Beneficiaries were divided into $3$ groups, each adding to the current standard of care. 
The first group exercised ARMMAN's current standard of care (CSOC) without additional ARMMAN-initiated calls.
In the second, the RMAB group,  ARMMAN staff added to the CSOC by initiating service calls to $225$ beneficiaries on average per week chosen by RMAB. The third was the Round-Robin group, where the exact same number of beneficiaries as the RMAB group were called every week based on a systematic sequential basis. 

\textit{Results from our study demonstrate that RMAB provides statistically significant improvement over CSOC and round-robin groups.} This improvement is also practically significant --- the RMAB group achieves a {$\sim30\%$} reduction in engagement drops over the other groups. Moreover, the round-robin group does not achieve statistically significant improvement over the CSOC group, i.e., RMAB's optimization of service calls is crucial. To the best of our knowledge, this is the first large-scale empirical validation of use of RMABs in a public health context. Based on these results, the RMAB system is currently being transitioned to ARMMAN to optimize service calls to their ever growing set of beneficiaries. Additionally, this methodology can be useful in assisting engagement in many other awareness or adherence programs, e.g., \citet{thirumurthy2012m, chen2021caries}. \textit{Our RMAB code would be released upon acceptance}.

\section{Related Work}

Patient adherence monitoring in healthcare has been shown to be an important problem \cite{martin2005challenge}, and is closely related to the churn prediction problem, studied extensively in the context of industries like telecom \cite{dahiya2015customer}, finance \cite{XIE20095445, shaaban2012proposed}, etc.
The healthcare domain has seen several studies on patient adherence for diseases like HIV \cite{HIV}, cardiac problems \cite{son2010application, corotto2013heart}, Tuberculosis \cite{Killian_2019, 10.1001/archinte.1996.00440020063008}, etc. These studies use a combination of patient background information and past adherence data, and build machine learning models to predict future adherence to prescribed medication
\footnote{Similarly, in our previous preliminary study (anonymous 2020) published in a non-archival setting, we used demographic and message features to build models for predicting beneficiaries likely to drop-off from ARMMAN's information program.}. However, such models treat adherence monitoring as a single-shot problem and are unable to appropriately handle the sequential resource allocation problem at hand. 
Additionally, the pool of beneficiaries flagged as high risk can itself be large, and the model can't be used to prioritize calls on a periodic basis, as required in our settings.

The Restless Multi-Armed Bandit (RMAB) framework has been popularly adopted to tackle such sequential resource allocation problems \cite{whittle-rbs, jung2019regret}. 
Computing the optimal solution for RMAB problems is shown to be PSPACE-hard. Whittle proposed an index-based heuristic \cite{whittle-rbs}, that can be solved in polynomial time and is now the dominant technique used for solving RMABs. It has been shown to be asymptotically optimal for the time average reward problem \cite{weber1990index}, and other families of RMABs arising from stochastic scheduling problems \cite{glazebrook2006indexable}. Several works as listed in Section~\ref{sec:intro}, show applicability of RMABs in different domains 
but these unrealistically assume perfect knowledge of the RMAB parameters, and have not been tested in real-world contexts. \citet{biswas2021learn, avrachenkov2020whittle}, present a Whittle Index-based Q-learning approach for unknown RMAB parameters. However, their techniques either assume identical arms or rely on receiving thousands of samples from each arm, which is unrealistic in our setting, given limited overall stay of a beneficiary in an information program --- a beneficiary may drop out or stop engaging with the program few weeks post enrolment unless a service call convinces them to do otherwise. Instead, we present a novel approach that applies clustering to the available historical data to infer model parameters.

Clustering in the context of Multi-Armed Bandit and Contextual Bandits has received significant attention in the past \cite{gentile2014online,li2019improved, yang2020exploring,li2021unifying}, but these settings do not consider restless bandit problems. 

\section{Preliminaries}
\label{sec:preliminaries}

\subsection{Background: Restless Multi-Armed Bandits}

An RMAB instance consists of $N$ independent 2-action Markov Decision Processes (MDP) \cite{mdp-puterman}, where each MDP is defined by the tuple $\{\mathcal{S}, \mathcal{A}, R, \mathcal{P}\}$. $\mathcal{S}$ denotes the state space, $\mathcal{A}$ is the set of possible actions, $R$ is the reward function $R: \mathcal{S} \times \mathcal{A} \times \mathcal{S} \rightarrow \mathbb{R}$ and $\mathcal{P}$ represents the transition function. We use $P_{s, s'}^{\alpha}$ to denote the probability of transitioning from state $s$ to state $s'$ under the action $\alpha$. The policy $\pi$, is a mapping $\pi : \mathcal{S} \rightarrow \mathcal{A}$ that selects the action to be taken at a given state.
The total reward accrued can be measured using either the discounted or average reward criteria to sum up the immediate rewards accrued by the MDP at each time step. Our formulation is amenable to both, although we use the discounted reward criterion in our study. 

The expected \emph{discounted reward} starting from state $s_0$ is defined as $V_{\beta}^\pi(s_0) = \mathbb{E}\left[\sum_{t=0}^\infty \beta^tR(s_t, \pi (s_t), s_{t+1}|\pi, s_0)\right]$ where the next state is drawn according to $s_{t+1} \sim P_{s_t,s_{t+1}}^{\pi(s_t)}$,  $\beta \in [0,1)$ is the discount factor and actions are selected according to the policy mapping $\pi$. 
The planner's goal is to maximize the total reward. 

\begin{figure}[htb]
\centering
\resizebox{!}{64pt}{%
\begin{tikzpicture}[
    -Triangle, every loop/.append style = {-Triangle},
    start chain=main going right,
    state/.style={circle,minimum size=9mm,draw},
      node distance=14mm,
      font=\scriptsize,
      >=stealth,
      bend angle=48,
      auto
    ]
  \node [state, on chain, fill=black, text=white] (A) {\textbf{\boldmath NE}};
   
  \node[state, on chain, fill=lightgray]  (B) {E};
 
    \path[->,draw, thick, bend right=45]
    (B) edge node[above] {$1-P_{E,E}^\alpha$} (A);
    \path[thick]
    (B) edge[loop right] node{$P_{E,E}^\alpha$} (B);
    
     \path[->,draw, thick, bend right=45]
    (A) edge node[below] {$P_{NE,E}^\alpha$} (B);
     \path[thick]
    (A) edge[loop left] node{$1-P_{NE,E}^\alpha$} (A);

\end{tikzpicture}
}

\caption{The beneficiary transitions from a current state $s$ to a next state $s'$ under action $\alpha$, with probability $P_{ss'}^\alpha$.}
\label{fig:MDP}
\end{figure}
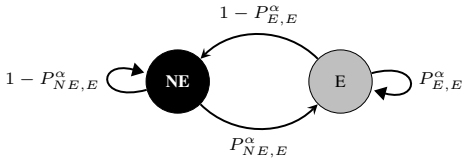

We model the engagement behavior of each beneficiary by an MDP corresponding to an arm of the RMAB. Pulling an arm corresponds to an active action, i.e., making a service call (denoted by $\alpha=a$), while $\alpha=p$ denotes the passive action of abstaining from a call. The state space $\mathcal{S}$ consists of binary valued states, $s$, that account for the recent engagement behavior of the beneficiary; $s \in [NE,E]$ (or equivalently, $s \in [0,1]$) where $E$ and $NE$ denote the `Engaging' and `Not Engaging' states respectively. For example, in our domain, ARMMAN considers that if a beneficiary stays on the automated voice message for more than 30 seconds (average message length is 1 minute), then the beneficiary has engaged. If a beneficiary engages at least once with the automated voice messages sent during a week, they are assigned the engaging ($E$) state for that time step and non-engaging ($NE$) state otherwise.
For each action $\alpha \in \mathcal{A}$, the beneficiary states follow a Markov chain represented by the 2-state Gilbert-Elliot model \cite{gilbert1960capacity} with transition parameters given by $P_{ss'}^\alpha$, as shown in Figure~\ref{fig:MDP}. With slight abuse of notation, the reward function $R(.)$ of $n^{th}$ MDP is simply given by $R_n(s)=s$ for $s \in \{0,1\}$.

We adopt the Whittle solution approach
described previously for solving the RMAB. It hinges around the key idea of a ``passive subsidy'', which is a hypothetical reward offered to the planner, in addition to the original reward function for choosing the passive action. 
The Whittle Index is then defined as the infimum subsidy that makes the planner indifferent between the `active' and the `passive' actions, i.e.,:
\begin{equation}
W(s)=inf_{\lambda}\{\lambda:Q_{\lambda}(s,p)=Q_{\lambda}(s,a)\}    
\end{equation}

\subsection{Data Collected by ARMMAN}
Beneficiaries enroll into ARMMAN's information program 
with the help of health workers, who collect the beneficiary's demographic data such as age, education level, income bracket, phone owner in the family, gestation age, number of children, preferred language and preferred slots for the automated voice messages during enrolment. These features are referred to as Beneficiary Registration Features in rest of the paper. 
Beneficiaries provided both written and digital consent for 
receiving automated voice messages and service calls.
ARMMAN also stores listenership information regarding the automated voice messages together with the registration data in an anonymized fashion.  




\section{Problem Statement}

We assume the planner has access to an offline historical data set of beneficiaries, $\mathcal{D}_{train}$. Each beneficiary data point $\mathcal{D}_{train}[i]$ consists of a tuple, $\langle f, \mathcal{E} \rangle$, where $f$ is beneficiary $i$'s  feature vector of static features,
and $\mathcal{E}$ is an episode storing the trajectory of $(s, \alpha, s')$ pairs for that beneficiary, where $s$ denotes the start state, $\alpha$ denotes the action taken (passive v/s active), and $s’$ denotes the next state that the beneficiary lands in after executing $\alpha$ in state $s$. We assume that these $(s,\alpha,s')$ samples are drawn according to fixed, latent transition matrices $P_{ss'}^a[i]$ and $P_{ss'}^p[i]$ (corresponding to the active and passive actions respectively), unknown to the planner, and potentially unique to each beneficiary. 

Given $D_{train}$, we now consider a new beneficiary cohort $\mathcal{D}_{test}$, consisting of $N$ beneficiaries, marked $\{1,2,\dots, N\}$, that the planner must plan service calls for. The MDP transition parameters corresponding to beneficiaries in $\mathcal{D}_{test}$ are unknown to the planner, but assumed to be drawn at random from a distribution similar to the joint distribution of features and transition parameters of beneficiaries in the historical data distribution. We assume the planner has access to the feature vector $f$ for each beneficiary in $\mathcal{D}_{test}$.

We now define the service call planning problem as follows. The planner has upto $m$ resources available per round, which the planner may spend towards delivering service calls to beneficiaries. Beneficiaries are represented by $N$ arms of the RMAB, of which the planner may pull upto $m$ arms (i.e., $m$ service calls) at each time step. We consider a round or timestep of one week which allows planning based on the most recent engagement patterns of the beneficiaries.



\section{Methodology}
\label{sec:method}

\begin{figure}[t]
    \centering
    \includegraphics[width=0.9\linewidth]{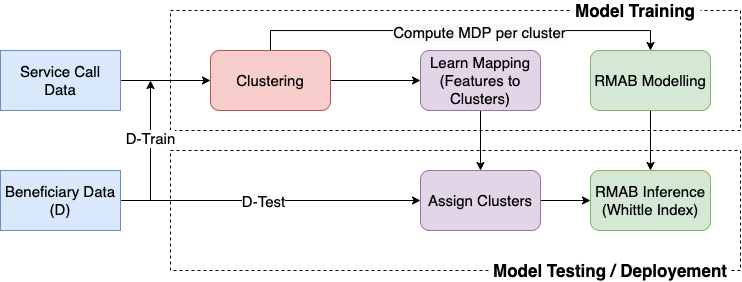}
    \caption{RMAB Training and Testing pipelines proposed}
    \label{fig:pipeline}
\end{figure}

Figure~\ref{fig:pipeline} shows our overall solution methodology.
We use clustering techniques that exploit historical data $D_{train}$ to estimate an offline RMAB problem instance relying solely on the beneficiaries' static features and state transition data. This enables overcoming the challenge of limited samples (time-steps) per beneficiary.
Based on this estimation, we use the Whittle Index approach to prioritize service calls. 

\subsection{Clustering Methods}

We use historical data $\mathcal{D}_{train}$ to learn the impact of service calls on transition probabilities. While there is limited service call data (active transition samples) for any single beneficiary, clustering on the beneficiaries allows us to combine their data to infer transition probabilities for the entire group.
Clustering offers the added advantage of reducing computational cost for resource limited NGOs; since all beneficiaries within a cluster share identical transition probability values we can compute their Whittle index all at once. 
We present four such clustering techniques below:



\paragraph{1. Features-only Clustering (FO):}

This method relies on the correlation between the beneficiary feature vector $f$ and their corresponding engagement behavior. We employ k-means clustering on the feature vector $f$ of all beneficiaries in the historic dataset $D_{train}$, and then derive the representative transition probabilities for each cluster by pooling all the $(s, \alpha, s’)$ tuples of beneficiaries assigned to that cluster. At test time, the features $f$ of a new, previously unseen beneficiary in $D_{test}$ map the beneficiary to their corresponding cluster and estimated transition probabilities. 

\paragraph{2. Feature + All Probabilities (FAP)}

In this 2-level hierarchical clustering technique, the first level uses a rule-based method, using features to divide beneficiaries into a large number of pre-defined buckets, $B$. Transition probabilities are then computed by pooling the $(s, \alpha, s’)$ samples from all the beneficiaries in each bucket. Finally, we perform a k-means clustering on the transition probabilities of these $B$ buckets to reduce them to $k$ clusters ($k\ll B$).
However, this method suffers from several smaller buckets missing or having very few active transition samples. 

\paragraph{3. Feature + Passive Probabilities (FPP):}

This method builds on the FAP method, but only considers the passive action probabilities to preclude the issue of  missing active transition samples.

\subsubsection{4. Passive Transition-Probability based Clustering (PPF):}

The key motivation here is 
to group together beneficiaries with similar transition behaviors, irrespective of their features. To this end, we use k-means clustering on passive transition probabilities (to avoid issues with missing active data) of beneficiaries in $D_{train}$ and identify cluster centers. 
We then learn a map $\phi$ from the feature vector $f$ to the cluster assignment of the beneficiaries that can be used to infer the cluster assignments of new beneficiaries at test-time solely from $f$.
We use a random forest model as 
$\phi$.

The rule-based clustering on features involved in \textit{FPP} and \textit{FAP} methods can be thought of as using one specific, hand-tuned mapping function $\phi$. In contrast, the \textit{PPF} method \textit{learns} such a map $\phi$ from data, eliminating the need to manually define accurate and reliable feature buckets. 



\subsection{Evaluation of Clustering Methods}
\label{subsec:comapre_clusters}
We use a historical dataset, $\mathcal{D}_{train}$ from ARMMAN consisting of 4238
beneficiaries in total, who enrolled into the program between May-July 2020.
We compare the clustering methods empirically, based on the criteria described below. 


\begin{figure*}[ht]
\centering
\begin{subfigure}[t]{0.23\textwidth}
    \centering
    \includegraphics[width=0.98\linewidth]{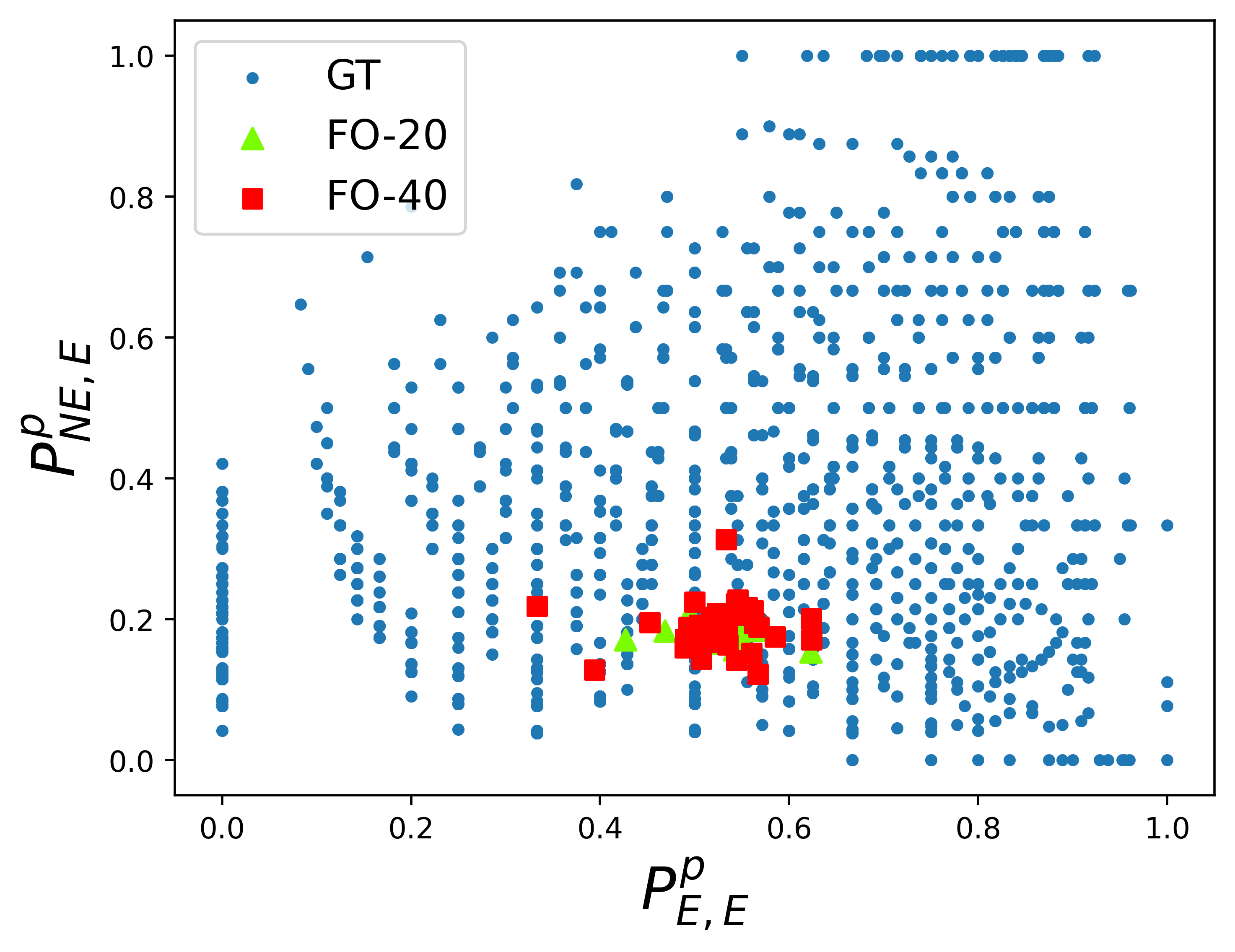}
    \caption{FO clustering}
    \label{fig:fo}
\end{subfigure}
\hfill
\begin{subfigure}[t]{0.23\textwidth}
    \centering
    \includegraphics[width=0.98\linewidth]{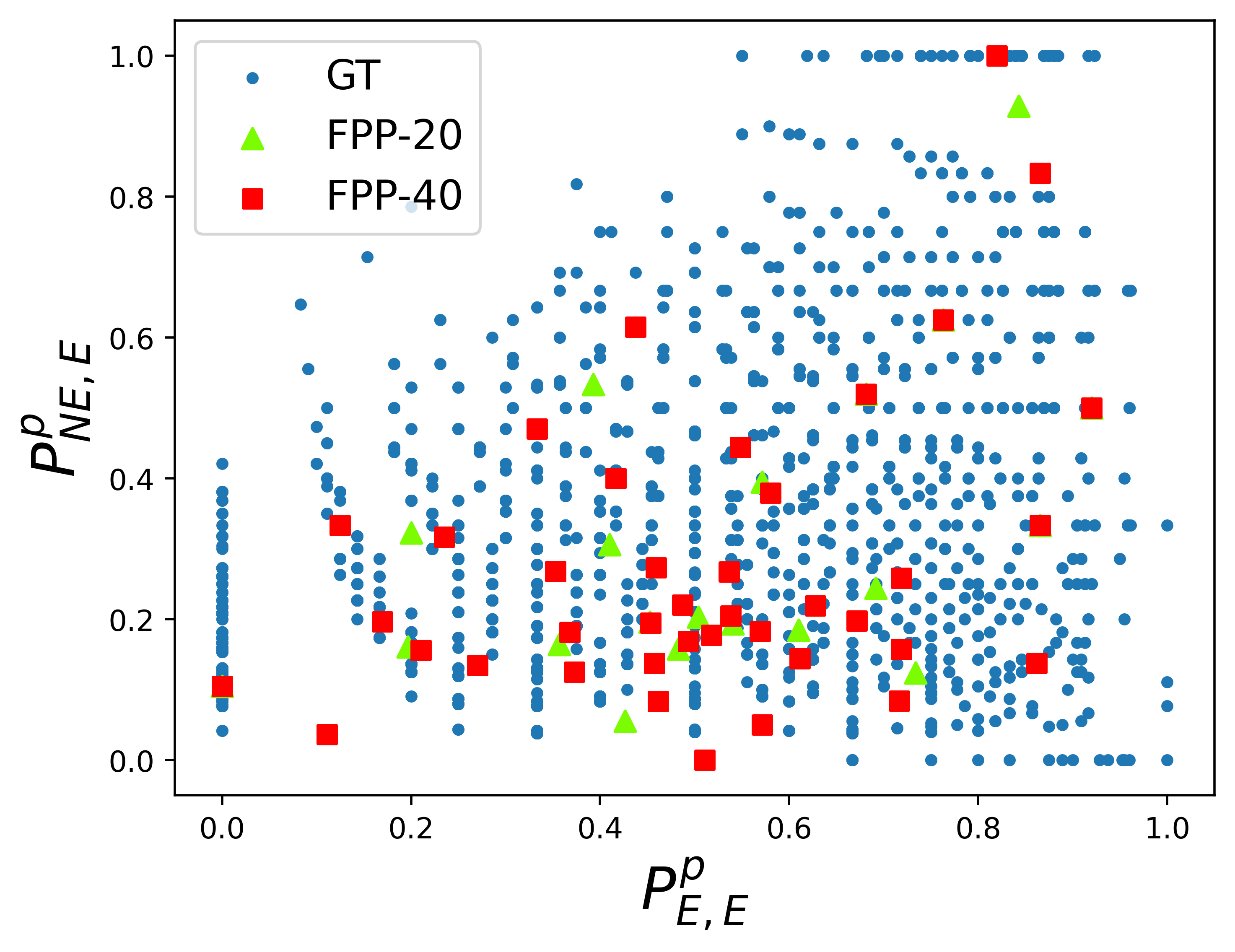}
    \caption{FPP clustering}
    \label{fig:fpp}
\end{subfigure}
\hfill
\begin{subfigure}[t]{0.23\textwidth}
    \centering
    \includegraphics[width=0.98\linewidth]{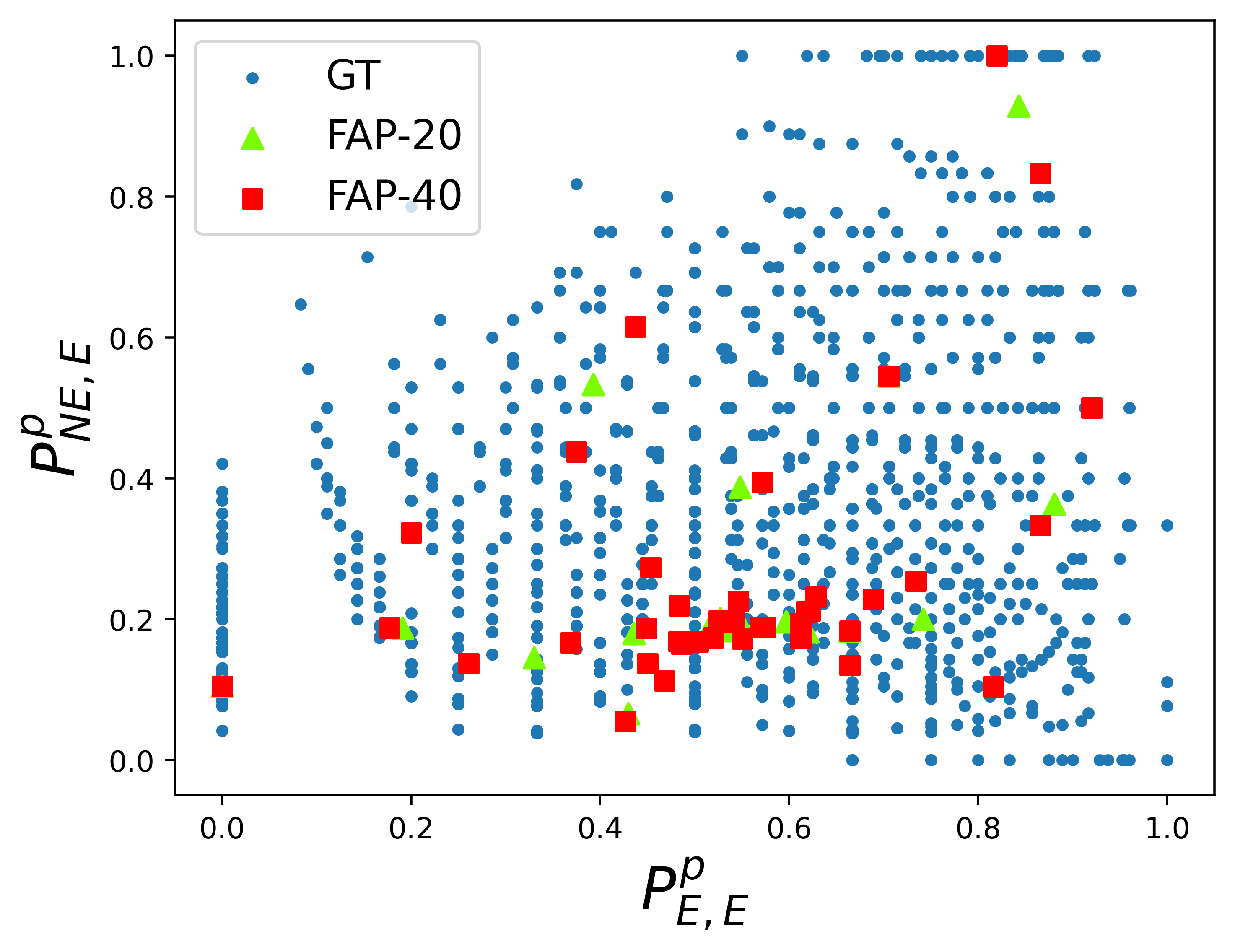}
    \caption{FAP clustering}
    \label{fig:fap}
\end{subfigure}
\hfill
\begin{subfigure}[t]{0.23\textwidth}
    \centering
    \includegraphics[width=0.98\linewidth]{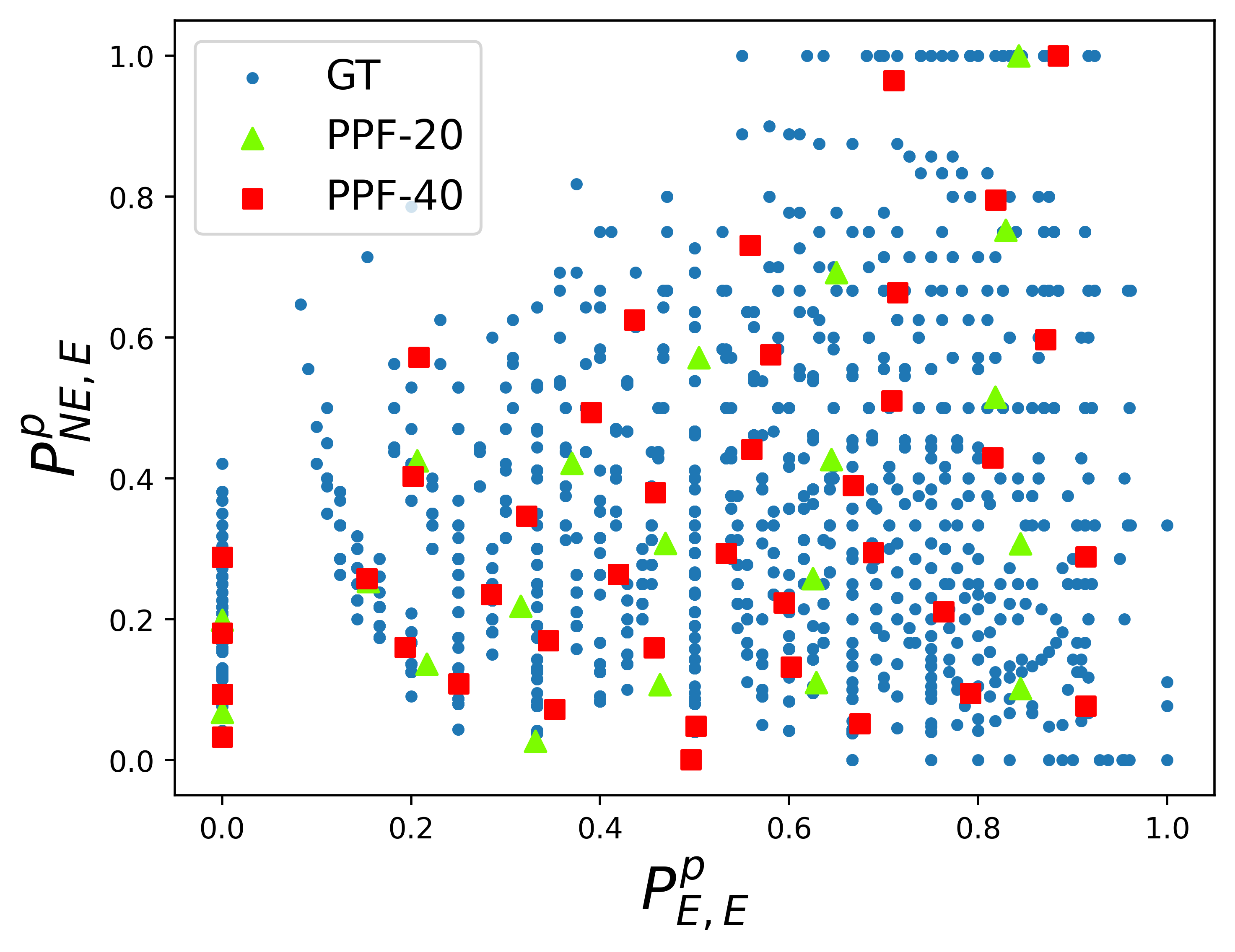}
    \caption{PPF clustering }
    \label{fig:ppf}
\end{subfigure}
\caption{Comparison of passive transition probabilities obtained from different clustering methods with cluster sizes $k=\{20,40\}$ with the ground truth transition probabilities. Blue dots represent the true passive transition probabilities for every beneficiary while red or green dots represent estimated cluster centres. }
\label{fig:clustering}
\end{figure*}

\textbf{1. Representation:} Cluster centers that are representative of the underlying data distribution better resemble the ground truth transition probabilities. This is of prime importance to the planner, who must rely on these values to plan actions. Fig~\ref{fig:clustering} plots the ground truth transition probabilities and the resulting cluster centers determined using the proposed methods. Visual inspection reveals that the \textit{PPF} method represents the ground truth well, as is corroborated by the quantitative metrics of Table \ref{tab:weekly-combined} that compares the RMSE error across different clustering methods.

\textbf{2. Balanced cluster sizes:} A low imbalance across cluster sizes is desirable to preclude the possibility of arriving at few, gigantic clusters which will assign identical whittle indices to a large groups of beneficiaries. Working with smaller clusters also aggravates the missing data problem in estimation of active transition probabilities. 
Considering the variance in cluster sizes and RMSE error for the different clustering methods with $k=\{20, 40\}$ as shown in 
Table \ref{tab:weekly-combined},  $PPF$ outperforms the other clustering methods and was chosen for the pilot study.

\begin{table}[h]
\centering
\caption{Average RMSE and cluster size variance over all beneficiaries for different methods. Total Beneficiaries = 4238, $\mu_{20} = 211.9$, $\mu_{40} = 105.95$ ($\mu = $ average beneficiaries per cluster)}
\begin{tabular}{|c|c|c|c|c|}
\hline
\multirow{2}{5em}{Clustering Method} & \multicolumn{2}{c|}{Average RMSE} & \multicolumn{2}{c|}{Standard Deviation} \\ \cline{2-5} 
                                   & \textbf{k = 20} & \textbf{k = 40} & \textbf{k = 20}    & \textbf{k = 40}    \\ \hline
\textbf{FO}                        & 0.229           & 0.228           & 143.30             & 74.22              \\ \hline
\textbf{FPP}                       & 0.223           & 0.222           & 596.19             & 295.01             \\ \hline
\textbf{FAP}                       & 0.224           & 0.223           & 318.46             & 218.37             \\ \hline
\textbf{PPF}                       & 0.041           & 0.027           & 145.59             & 77.50              \\ \hline
\end{tabular}
\label{tab:weekly-combined}
\end{table}

Next we turn to choosing $k$, the number of clusters: as $k$ grows, the clusters become sparse in number of active samples aggravating the missing data problem while a smaller $k$ suffers from a higher RMSE. We found $k=40$ to be optimal and chose it for the pilot study
. 

Finally, we adopt the Whittle solution approach for RMABs to plan actions and pre-compute all of the possible $2*k$ index values that beneficiaries can take (corresponding to combinations of $k$ possible clusters and $2$ states). The indices can then be looked up at all future time steps in constant time, making this an optimal solution for large scale deployment with limited compute resources.

As we got this RMAB system ready for real-world use, there was
as an important observation for social impact settings: real-world use also required us to carefully handle several domain specific challenges, which were time consuming. For example, despite careful clustering,  a few clusters may still be missing active probability values, which required employing a data imputation heuristic 
. Moreover, there were other constraints specific to ARMMAN, such as a beneficiary should receive only one service call every $\eta$ weeks, which was addressed by introducing ``sleeping states'' for beneficiaries who receive a service call 
.





\section{Experimental Study}


In this section, we discuss a real-world quality improvement study. 
We also simulate the expected outcome in other synthetically constructed situations and demonstrate good performance of our approach across the board. 


\subsection{Service Quality Improvement Study}

\subsubsection{Setup}
This cohort of beneficiaries registered in the program between Feb 16, 2021 and March 15, 2021 as $D_{test}$ and started receiving automated voice messages few days post enrolment as per their gestational age. Additionally, as per the current standard of care, any of these beneficiaries could initiate a service call by placing a ``missed call''. 
The $23003$ beneficiaries are randomly distributed across 3 groups, each group adding to the CSOC as follows:

\begin{itemize}

    \item \textbf{Current-Standard-of-Care (CSOC) Group}: The beneficiaries in this group follow the original standard of care, where there are no ARMMAN initiated service calls. The listenership behavior of beneficiaries in this group is used as a benchmark for the RR and RMAB groups. 
    \item \textbf{RMAB group}: In this group, beneficiaries are selected for ARMMAN-initiated service call per week via the Whittle Index policy described in Section \ref{sec:preliminaries}. Even though all beneficiaries within a cluster are modeled by identical MDP parameters, their states may evolve independently, and so the Whittle indices are tracked for each beneficiary separately, leading to an RMAB with $7668$ arms.
    \item \textbf{Round Robin (RR) group}:  By default, NGOs including ARMMAN  often conduct service calls using some systematic set order  -- the idea here is to have an easily executable policy, that services enough of a cross-section of beneficiaries and can be scaled up or down per week based on available resources. To recreate this setting, we generate service calls to beneficiaries based on the ascending order of their date of enrollment for this RR group, as recommended by ARMMAN. If this method succeeds compared to CSOC, then a simple manual strategy is enough; RMAB style optimization may not be needed.
    
\end{itemize}

. Table \ref{tab:initial} shows the absolute number of beneficiaries in states E or NE, where the state is computed using one week of engagement data between April 19 - April 26, 2021. 


Beneficiaries across all three groups receive the same automated voice messages regarding pregnancy and post-birth care throughout the program, and no health related information is withheld from any beneficiary. The study only aims to evaluate the effectiveness of ARMMAN-initiated outbound service calls with respect to improving engagement with the program across the three groups. No interviews or research data or feedback was collected from the beneficiaries.

\begin{table}[t]
\centering
\caption{Beneficiary distribution in the three groups and their start states during week 0 of the study.}
\begin{tabular}{|c|c|c|c|}
\hline
\multirow{2}{3em}{\textbf{Group}} & \multirow{1}{4em}{\textbf{Engaging}} & \multirow{1}{6.5em}{\textbf{Non-Engaging}} & \multirow{2}{2em}{\textbf{Total}}\\ 
& \textbf{(E)} & \textbf{(NE)} & \\\hline
RMAB & 3571 & 4097 & 7668\\ \hline
RR & 3647 & 4021 & 7668\\ \hline
CSOC & 3661 & 4006 & 7667\\ \hline
\end{tabular}
\label{tab:initial}
\end{table}

The study started on April 26, 2021, with $m$ beneficiaries  selected from the RMAB and RR group each ($m\ll N$) per week for ARMMAN-initiated service calls.  ARMMAN staff performing service calls were blind to the experimental groups that the beneficiaries belonged to. Recall, the goal of the service calls is to encourage the beneficiaries to engage with the health information message program in the future. For this study, number of service calls $m$ was on average $225$ per week for each of RMAB and RR groups to reflect real-world constraints on service calls. The study was scheduled for a total of 7 weeks, during which $20\%$ of the RMAB (and RR) group had received a service call, which is closer to the percentage of population that may be reached in service calls by ARMMAN. 
 \footnote{Each beneficiary group also received very similar beneficiary-initiated calls, but these were less than 10\% of the ARMMAN-initiated calls in RMAB or RR groups over 7 weeks.}
 

\subsubsection{Results}
\begin{figure}[htb]
    \centering
    \includegraphics[width=0.8\columnwidth]{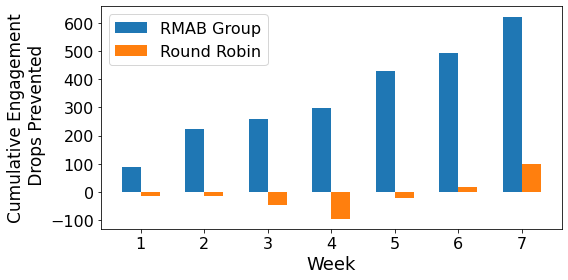}
    \caption{Cumulative number of weekly engagement drops prevented (in comparison to the CSOC group) by RMAB far exceed those prevented by RR.}
    \label{fig:engagement-drop-prevented}
\end{figure}

We present our key results from the study in Figure \ref{fig:engagement-drop-prevented}. The results are computed at the end of 7 weeks from the start of the quality improvement study on April 26, 2021.


Figure \ref{fig:engagement-drop-prevented} measures the impact of service calls by the RMAB and RR policies in comparison to the CSOC Group. Beneficiaries' engagement with the program typically starts to dwindle with time. 
In Figure \ref{fig:engagement-drop-prevented}, we measure the impact of a service call policy as the cumulative drop in engagement prevented compared to the CSOC Group. We consider  drop in engagement instead of the raw engagement numbers themselves, because of the slight difference in the numbers of beneficiaries in engaging (E) state at the start of the study. 
The drop in engagement under a policy $\pi$ at time $t$ can be measured as the change in engagement: 
\begin{equation}
\Delta_{current}^{\pi}(t) \coloneqq \sum\limits_{n \in N} (R_n(s_0) - R_n(s_t))
\end{equation}
where $R_n(s_t)$ represents the reward for $n^{th}$ beneficiary in state $s_t$ at time step $t$ and cumulative drop in engagement is:
\begin{equation}
    \Delta_{cumulative}^{\pi}(t) \coloneqq \sum\limits_{n \in N} \sum\limits_{\zeta=0}^{\zeta=t} (R_n(s_0) - R_n(s_\zeta))
\end{equation}
The cumulative drop in engagement prevented by a policy $\pi$, in comparison to the CSOC Group is thus simply:
\begin{equation}
    \Delta_{cumulative}^{\pi}(t)- \Delta_{cumulative}^{CSOC}(t)
\end{equation}
and is plotted on the y-axis of Figure~\ref{fig:engagement-drop-prevented}. 

Figure~\ref{fig:engagement-drop-prevented} shows that the RMAB policy prevents a total 622 instances of a drop in automated health message engagement, at the end of 7 weeks, as compared to CSOC. RR group, on the other hand, only prevents 101 engagement drops by the end of week 7. Given that there are a total of 1944 engagement drops in the CSOC group, we show in the first row of Table \ref{tab:stat-sig}, that the RMAB group has 32.0\% and 28.3\% less cumulative engagement drops as compared to the CSOC and RR groups respectively by the end of the study.

\subsubsection{Statistical Analysis}


\begin{table}[h]
\centering
\caption{Statistical significance for service call policy impact at week 7 is tested using a linear regression model. We use: $^*p < 0.05;\; ^{\dagger}p < 0.1$}
\begin{tabular}{|p{9.3em}|p{3.9em}|p{3.3em}|p{3.3em}|}
\hline
                                         & \textbf{RMAB vs CSOC} & \textbf{RR vs CSOC} & \textbf{RMAB vs RR} \\ \hline
\% reduction in cumulative engagement drops & \multirow{2}{*}{32.0\%} &  \multirow{2}{*}{5.2\%}    & \multirow{2}{*}{28.3\%}   \\ \hline

p-value                                      & 0.044$^*$             & 0.740                    & 0.098$^\dagger$           \\ \hline

Coefficient $\beta$                         & -0.0819        &-0.0137           &-0.0068                         \\ \hline

\end{tabular}
\label{tab:stat-sig}
\end{table}

To investigate the benefit from use of RMAB policy over policies in the RR and CSOC groups, we use regression analysis \cite{angrist2008mostly}. Specifically, we fit a linear regression model to predict number of cumulative engagement drops at week 7 while controlling for treatment assignment and covariates specified by beneficiary registration features.
The model is given by:
$$Y_i = k + \beta{T_i} + \sum\limits_{j=1}^{J}\gamma_jx_{ij} + \epsilon_i$$
where for the $i_{th}$ beneficiary, $Y_i$ is the outcome variable defined as number of cumulative engagement drops at week 7, $k$ is the constant term, $\beta$ is the treatment effect, $T_i$ is the treatment indicator variable, $x_i$ is a vector of length $J$ representing the $i_{th}$ beneficiary's registration features, $\gamma_j$ represents the impact of the $j_{th}$ feature on the outcome variable and $\epsilon_i$ is the error term. 
For evaluating the effect of RMAB service calls as compared to CSOC group, we fit the regression model only for the subset of beneficiaries assigned to either of these two groups. $T_i$ is set to $1$ for beneficiaries belonging to the RMAB group and $0$ for those in CSOC group. We repeat the same experiment to compare RR vs CSOC group and RMAB vs RR group. 

The results are summarized in Table \ref{tab:stat-sig}. 
We find that RMAB has a statistically significant treatment effect in reducing cumulative engagement drop (negative $\beta, p < 0.05$) as compared to CSOC group. However,  the treatment effect is not statistically significant when comparing RR with CSOC group ($p = 0.740$).
Additionally, comparing RMAB group with RR, we find $\beta$, the RMAB treatment effect, to be significant $(p<0.1)$. 
This shows that RMAB policy has a statistically significant
effect on reducing cumulative engagement drop
as compared to both the RR policy and CSOC. RR fails to achieve statistical significance against CSOC. Together these results illustrate the importance of RMAB’s optimization of service calls, and that without such optimization, service calls may not yield any benefits.

\subsubsection{RMAB Strategies}

\begin{figure}[t]
\centering
\begin{subfigure}[b]{0.45\columnwidth}
    \centering
    \includegraphics[width=0.98\linewidth]{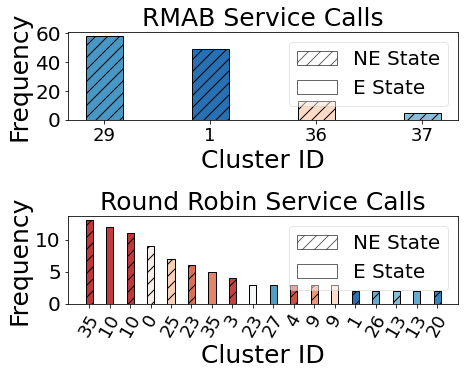}
    \caption{Week 1 Service Calls}
    \label{fig:hue-week1}
\end{subfigure}
\hfill
\begin{subfigure}[b]{0.54\columnwidth}
    \centering
    \includegraphics[width=0.98\linewidth]{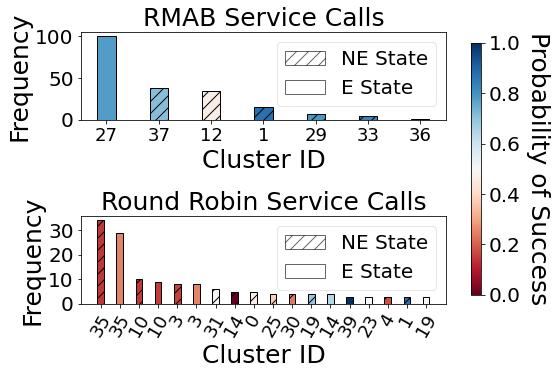}
    \caption{Week 2 Service Calls}
    \label{fig:hue-week2}
\end{subfigure}

\caption{
Distributions of clusters picked for service calls by RMAB and RR are significantly different. RMAB is very strategic in picking only a few clusters with a promising probability of success, RR displays no such selection.
}
\label{fig:hue-distribution}
\end{figure}

We analyse RMAB's strategic selection of beneficiaries in comparison to RR using Figure ~\ref{fig:hue-distribution}, where we group beneficiaries according to their whittle indices, equivalently their $\langle \texttt{cluster, state} \rangle$. 
Figure \ref{fig:hue-distribution} plots the frequency distribution of beneficiaries (shown via corresponding clusters) who were selected by RMAB and RR 
in the first two weeks. For example, the top plot in Figure~\ref{fig:hue-week1} shows that RMAB selected $60$ beneficiaries from cluster 29 (NE state). 
First, we observe that RMAB was clearly more selective, choosing beneficiaries from just four (Figure~\ref{fig:hue-week1}) or seven (Figure~\ref{fig:hue-week2}) clusters, rather than RR that chose from 20. Further, we assign each cluster a hue based on their probability of transitioning to engaging state from their current state given a service call. Figure \ref{fig:hue-distribution} reveals that RMAB consistently prioritizes clusters with high probability of success (blue hues) while RR deploys no such selection; its distribution emulates the overall distribution of beneficiaries across clusters (mixed blue and red hues).

\begin{figure}[t]
\centering
\begin{subfigure}[t]{0.45\columnwidth}
    \centering
    \includegraphics[width=0.95\columnwidth]{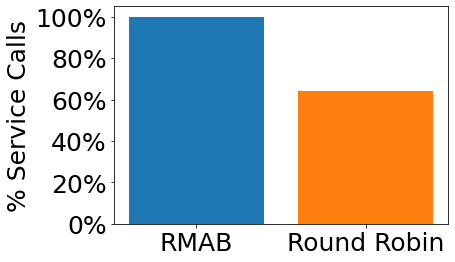}
    \caption{}
    \label{fig:resource-spent}
\end{subfigure}
\hfill
\begin{subfigure}[t]{0.45\columnwidth}
    \centering
    \includegraphics[width=0.95\columnwidth]{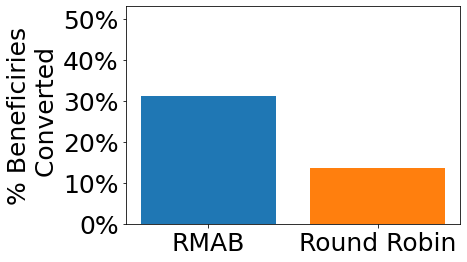}
    \caption{}
    \label{fig:conversion-rate}
\end{subfigure}
\caption{(a) \% of week 1 service calls on non-engaging beneficiaries (b) \% of non-engaging beneficiaries of week 1 receiving service calls that converted to engaging by week 7}
\label{fig:week1-conversion-rate-resource-spent}
\end{figure}

Furthermore, Figure \ref{fig:resource-spent} further highlights 
the situation in week $1$, where RMAB spent 100\% of its service calls on beneficiaries in the non-engaging state while RR spent the same on only 64\%.
Figure \ref{fig:conversion-rate} shows 
that RMAB converts $31.2\%$ of the beneficiaries shown in Figure \ref{fig:resource-spent} from non-engaging to engaging state by week 7, while RR does so for only 13.7\%.
This further illustrates the need for optimizing service calls for them to be effective, as done by RMAB.

\subsection{Synthetic Results}
We run additional simulations to test other service call policies beyond those included in the quality improvement study and confirm the superior performance of RMAB. 
Specifically, we compare to the following baselines: (1) \textsc{Random} is a naive baseline that selects $m$ arms at random. (2) \textsc{Myopic} is a greedy algorithm that pulls arms optimizing for the reward in the immediate next time step. \textsc{Whittle} is our algorithm.
We compute a normalized reward of an algorithm $\texttt{ALG}$ as:
$\frac{100\times (\overline{R}^{\texttt{ALG}}- \overline{R}^{\text{CSOC}})}{\overline{R}^{\text{WHITTLE}}- \overline{R}^{\text{CSOC}}}$
where $\overline{R}$ is the total discounted reward. Simulation results are averaged over 30 independent trials and run over 40 weeks.

\begin{figure}[t]
\centering
\includegraphics[width=0.85\linewidth]{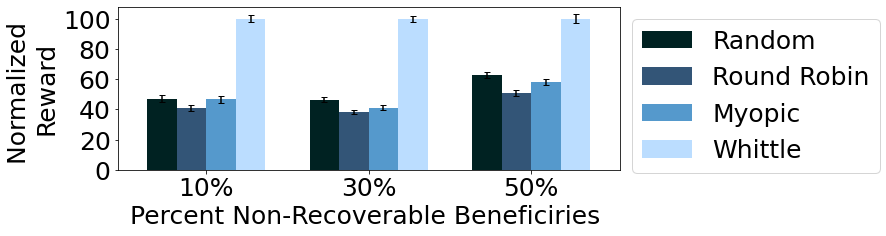}
\caption{Performance of \textsc{myopic} can be arbitrarily bad and even worse than \textsc{Random}, unlike the Whittle policy.}
\label{fig:counterExample}
\end{figure}

Figure~\ref{fig:counterExample} presents simulation of an adversarial example \cite{mate2020collapsing} consisting of 
$x\%$ of non-recoverable and $100-x\%$ of self-correcting beneficiaries for different values of $x$.
Self-correcting beneficiaries tend to miss automated voice messages sporadically, but revert to engaging ways without needing a service call. Non-recoverable beneficiaries are those who may drop out for good, if they stop engaging. We find that in such situations, MYOPIC proves brittle, as it performs even worse than RANDOM while WHITTLE performs well consistently. The actual quality improvement study cohort consists of $48.12\%$ non-recoverable beneficiaries (defined by $P_{01}^p<0.2$) and the remaining comprised of self-correcting and other types of beneficiaries. 




\section{Conclusions and Lessons Learned}
The widespread use of cell-phones, particularly in the global south, has enabled non-profits to launch massive programs delivering key health messages to a broad population of beneficiaries in a cost-effective manner. We present an RMAB based system to assist these non-profits in optimizing their limited service  resources. To the best of our knowledge,  ours is the first study to demonstrate the effectiveness of such RMAB-based resource optimization in  real-world public health contexts. These encouraging results have initiated the transition of our RMAB software to ARMMAN for real-world deployment. We hope this work paves the way for use of RMABs in many other health service applications.

Some key lessons learned from this research, which complement some of the lessons outlined in \cite{wilder2021clinical, floridi2020ai4sg, tomasev2020positive} include the following. First, social-impact driven engagement and design iterations with the NGOs on the ground is crucial to understanding the right AI model for use and appropriate research challenges. As discussed in footnote 1, our initial effort used a one-shot prediction model, and only after some design iterations we arrived at the current RMAB model. Next, given the missing parameters in RMAB, we found that the assumptions made in literature for learning such paramters did not apply in our domain, exposing new research challenges in RMABs. In short, \textit{domain partnerships with NGOs to achieve real social impact automatically revealed requirements for use of novel application of an AI model (RMAB) and new research problems in this model}. 


Second, \textit{data and compute limitations of non-profits are a real world constraint, and must be seen as genuine research challenges in AI for social impact, rather than  limitations.} In our domain, one key technical contribution in our RMAB system is deploying clustering methods on offline historical data to infer unknown RMAB parameters. Data is limited as not enough samples are available for any given beneficiary, who may stay in the program for a limited time. Non-profit partners also cannot bear the burden of massive compute requirements. 
Our clustering approach allows efficient offline mapping to Whittle indices, addressing both data and compute limits, enabling scale-up to service 10s if not 100s of thousands of beneficiaries. 
Third, \textit{in deploying AI systems for social impact, there are many technical challenges that may not need innovative solutions, but they are critical to deploying solutions at scale.} 
Indeed, deploying any system in the real world is challenging, but even more so in  domains where NGOs may be interacting with low-resource communities. 
Finally we hope this work serves as a useful example of deploying an AI based system for social impact in partnership with non-profits in the real world and will pave the way for more such solutions with real world impact.

\section*{Acknowledgement}
We thank Bryan Wilder, Aakriti Kumar for valuable feedback throughout the project and Divy Thakkar and Manoj Sandur Karnik for program management help. We also thank Suresh from ARMMAN team for helping set up the deployement pipeline. Additionally, we are grateful for support from the ARMMAN staff who made this field study possible.

\bibliography{ref}
\end{document}